\documentclass[runningheads]{llncs}
\usepackage{graphicx}
\graphicspath{{fig/}}

\usepackage{epsfig}
\usepackage{amsmath,amssymb} 
\usepackage{multirow}
\usepackage{booktabs}
\usepackage{diagbox}
\usepackage{tabularx}
\usepackage{xcolor}
\usepackage{nth}
\usepackage{adjustbox}

\usepackage[binary-units=true]{siunitx}
\DeclareSIUnit\fps{fps}

\usepackage{colortbl}
\definecolor{Gray}{gray}{0.85}

\usepackage{xspace}
\makeatletter
\DeclareRobustCommand\onedot{\futurelet\@let@token\@onedot}
\def\@onedot{\ifx\@let@token.\else.\null\fi\xspace}

\def\eg{\emph{e.g}\onedot} 
\def\ie{\emph{i.e}\onedot} 
 
\def\etc{\emph{etc}\onedot} \def\vs{\emph{vs}\onedot}
 
\def\etal{\emph{et al}\onedot}
\makeatother

\usepackage{hyperref}
\hypersetup{
        final,
        colorlinks,
        linkcolor={red!60!black},
        citecolor={blue!30!black},
        urlcolor={blue!90!black}
        }

\begin{document}
\pagestyle{headings}
\mainmatter
\def\ECCVSubNumber{3601}  

\title{Unified Image and Video Saliency Modeling}

%
\author{
Richard Droste\thanks{Richard Droste and Jianbo Jiao contributed equally to this work.} \and
Jianbo Jiao\textsuperscript{\thefootnote} \and
J.\ Alison Noble
}
%
\authorrunning{R. Droste et al.}
%
\institute{
University of Oxford\\
\email{\{richard.droste, jianbo.jiao, alison.noble\}@eng.ox.ac.uk}
}
\maketitle

\begin{abstract}
Visual saliency modeling for images and videos is treated as two independent tasks in recent computer vision literature.
While image saliency modeling is a well-studied problem and progress on benchmarks like \mbox{SALICON} and MIT300 is slowing, video saliency models have shown rapid gains on the recent DHF1K benchmark.
Here, we take a step back and ask: Can image and video saliency modeling be approached via a unified model, with mutual benefit?
We identify different sources of domain shift between image and video saliency data and between different video saliency datasets as a key challenge for effective joint modelling.
To address this we propose four novel domain adaptation techniques---
Domain-Adaptive Priors, Domain-Adaptive Fusion, Domain-Adaptive Smoothing and Bypass-RNN---
in addition to an improved formulation of learned Gaussian priors.
We integrate these techniques into a simple and lightweight encoder-RNN-decoder-style network, UNISAL, and train it jointly with image and video saliency data.
We evaluate our method on the video saliency datasets DHF1K, Hollywood-2 and UCF-Sports, and the image saliency datasets SALICON and MIT300.
With one set of parameters, UNISAL achieves state-of-the-art performance on all video saliency datasets and is on par with the state-of-the-art for image saliency datasets, despite faster runtime and a 5 to 20-fold smaller model size compared to all competing deep methods.
We provide retrospective analyses and ablation studies which confirm the importance of the domain shift modeling.
The code is available at \url{https://github.com/rdroste/unisal}.
\end{abstract}

\section{Introduction}
\label{sec:introduction}

When processing static scenes (images) and dynamic scenes (videos), humans direct their visual attention towards important information, which can be measured by recording eye fixations.
The task of predicting the fixation distribution is referred to as \emph{(visual) saliency prediction/modeling}, and the predicted distributions as \emph{saliency maps}.
Convolutional neural networks (CNNs) have emerged as the most performant technique for saliency modeling due to their capacity to learn complex feature hierarchies from large-scale datasets \cite{Borji2018,jiang2015salicon}.

While most prior work focuses on image data, interest in video saliency modeling was recently accelerated through ACLNet, a dynamic saliency model that outperforms static models on the large-scale, diverse DHF1K benchmark~\cite{Wang_2018_CVPR}.
However, as methods for video saliency modeling progress, it is usually considered a separate task to image saliency prediction~\cite{bak2017spatio,wang2019revisiting,Jiang2018,Min_2019_ICCV,Linardos2019,lai2019video} although both strive to model human visual attention.
Current dynamic models use image data only for pre-training \cite{bak2017spatio,Jiang2018,Min_2019_ICCV,Linardos2019,lai2019video} or auxiliary loss functions \cite{Wang_2018_CVPR}.
In addition, many dynamic models are incompatible with image inputs since they require optical flow \cite{bak2017spatio,lai2019video} or fixed-length video clips for spatio-temporal convolutions \cite{Jiang2018,Min_2019_ICCV}.
In this paper, we ask the question:
\emph{Is it possible to model static and dynamic saliency via one unified framework, with mutual benefit?}

\begin{figure}[t]
\centering
\includegraphics[width=0.6\textwidth]{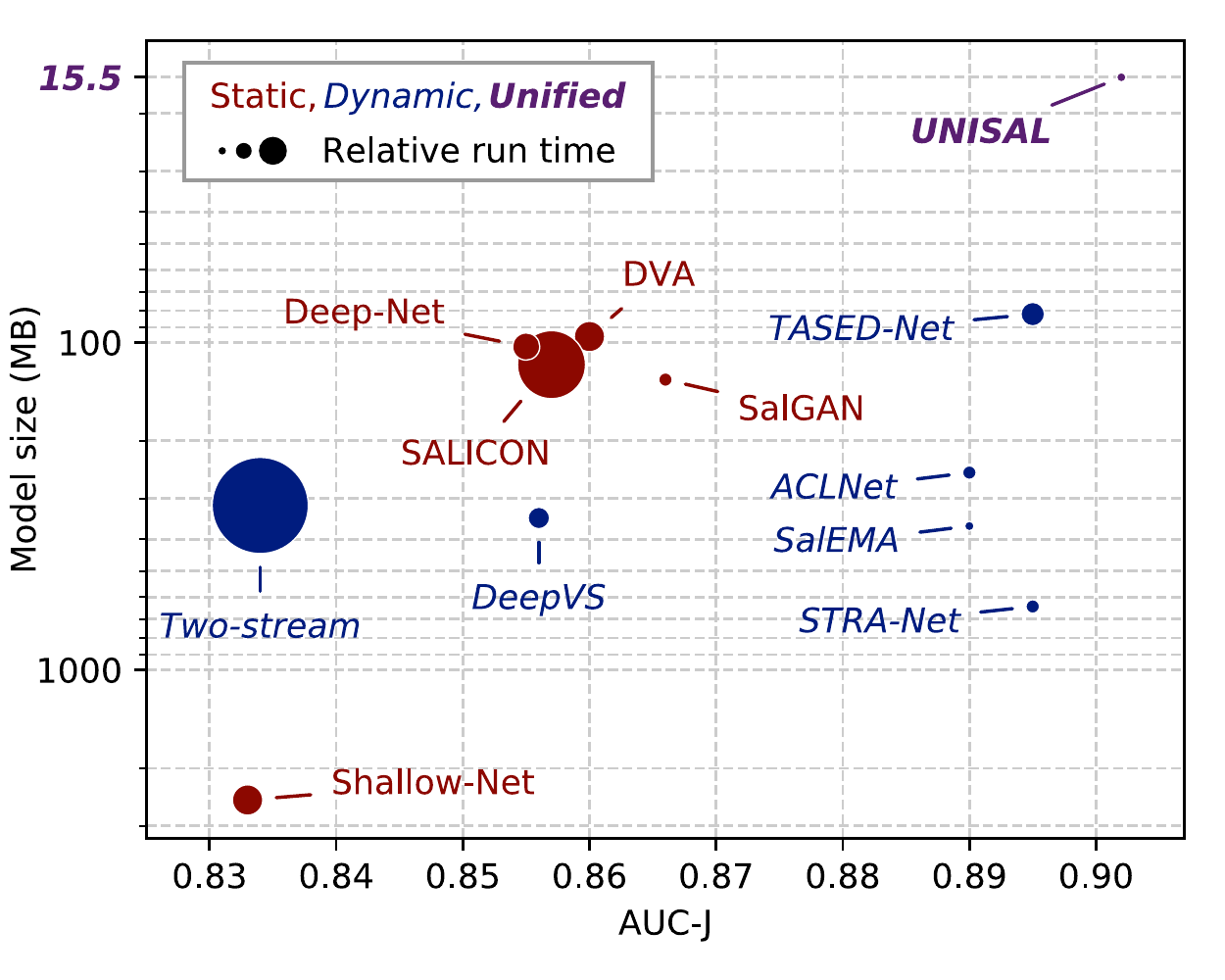}
\caption{
Comparison of the proposed model with current state-of-the-art methods on the DHF1K benchmark \cite{Wang_2018_CVPR}.
The proposed model is more accurate (as measured by the official ranking metric AUC-J \cite{Bylinskii2019}) despite a model size reduction of 81\% or more.
}
\label{fig:teaser}
\end{figure}

First, we present experiments that identify the domain shift between image and video saliency data and between different video saliency datasets as a crucial hurdle for joint modelling.
Consequently, we propose suitable domain adaptation techniques for the identified sources of domain shift.
To study the benefit of the proposed techniques, we introduce the UNISAL neural network, which is designed to model visual saliency on image and video data coequally while aiming for simplicity and low computational complexity.
The network is simultaneously trained on three video datasets---DHF1K~\cite{Wang_2018_CVPR}, Hollywood-2 and UCF-Sports~\cite{Mathe2015}---and one image saliency dataset, \mbox{SALICON}~\cite{jiang2015salicon}.

We evaluate our method on the four training datasets, among which DHF1K and SALICON have held-out test sets.
In addition, we evaluate on the established MIT300 image saliency benchmark \cite{Judd2012}.
We find that our model significantly outperforms current state-of-the-art methods on all video saliency datasets and achieves competitive performance for the image saliency datasets, with a fraction of the model size and faster runtime than competing models.
The performance of UNISAL on the challenging DHF1K benchmark is shown in Figure~\ref{fig:teaser}.
In summary, our contributions are as follows:
\begin{itemize}
\item To the best of our knowledge, we make the first attempt to model image and video visual saliency with one unified framework.
\item We identify different sources of domain shift as the main challenge for joint image and video saliency modeling and propose four novel domain adaptation techniques to enable strong shared features: Domain-Adaptive Priors, Domain-Adaptive Fusion, Domain-Adaptive Smoothing, and Bypass-RNN.
\item Our method achieves state-of-the-art performance on all video saliency datasets and is on par with the state-of-the-art for all image saliency datasets.
At the same time, the model achieves a 5 to 20-fold reduction in model size and faster runtime compared to all existing deep saliency models.
\end{itemize}

\section{Related Work}
\label{sec:related}
\subsubsection{Image Saliency Modeling.}
Most visual saliency modeling literature aims to predict human visual attention mechanisms on static scenes.
Early saliency models~\cite{itti1998model,borji2012state,sun2003object,harel2007graph,le2006coherent,Judd_2009} focus on low-level image features such as intensity/contrast, color, edges, \etc, and are are therefore referred to as \emph{bottom-up} methods.
Recently, the field has achieved significant performance gains through deep neural networks and their capacity to learn high-level, \emph{top-down} features, starting with Vig~\etal~\cite{vig2014large} who propose the first neural network-based approach.
Jiang~\etal~\cite{jiang2015salicon} collect a large-scale saliency dataset, SALICON, to facilitate the exploration of deep learning-based saliency modeling.
Zheng~\etal~\cite{zheng2018task} investigate the impact of high-level observer tasks on saliency modeling.
Other papers mainly focus on network architecture design with increasing model sizes.
For instance, Pan~\etal~\cite{pan2016shallow} evaluate shallow and deep CNNs for saliency prediction, and Kruthiventi~\etal~\cite{Kruthiventi2015} introduce dilated convolutions and Gaussian priors into the VGG network architecture.
Kuemmerer~\etal~\cite{Kummerer2016} propose a simplified VGG-based network while Wang~\etal~\cite{wang2017deep} add skip connections to fuse multiple scales and Cornia~\etal~\cite{Cornia2016a} add an attentive convolutional LSTM and learned Gaussian priors.
Yang~\etal~\cite{yang2019dilated} expand on the idea of dilated convolutions based on the inception network architecture.
While exploration is still ongoing for image saliency modeling, dynamic scenes are arguably at least as relevant to human visual experience, but have received less attention in the literature to date.

\subsubsection{Video Saliency Modeling.}
Similar to image saliency models, early dynamic models~\cite{marat2009modelling,mahadevan2009spatiotemporal,rudoy2013learning,hou2009dynamic} predict video saliency based on low-level visual statistics, with additional temporal features (\eg, optical flow).
Marat~\etal~\cite{marat2009modelling} use video frame pairs to compute a static and a dynamic saliency map, which are fused for the final prediction.
Marat~\etal~\cite{marat2009modelling} and Zhong~\etal~\cite{zhong2013video} combine spatial and temporal saliency features and fuse the predictions.
By extending the center-surround saliency in static scenes, Mahadevan~\etal~\cite{mahadevan2009spatiotemporal} use dynamic textures to model video saliency.
The performance of these early models is limited by the ability of the low-level features to represent temporal information.
Consequently, deep learning based methods have been introduced for dynamic saliency modeling in recent years.
Gorji~\etal~\cite{gorji2018going} propose to incorporate attentional push for video saliency prediction, via a multi-stream convolutional long short-term memory network (ConvLSTM).
Jiang~\etal~\cite{Jiang2018} show that human attention is attracted to moving objects and propose a saliency-structured ConvLSTM to generate video saliency.
A recent work~\cite{wang2019revisiting} presents a new large-scale video saliency dataset, DHF1K, and propose an attention mechanism with ConvLSTM to achieve better performance than static deep models.
The DHF1K dataset, sparked advances~\cite{Min_2019_ICCV,lai2019video,Linardos2019} in video saliency prediction, exploring different strategies to extract temporal features (optical flow, 3D convolutions, different recurrences).
However, the above methods either extend prior image saliency models or focus on video data alone with limited applicability to static scenes.
Guo~\etal~\cite{guo2008spatio} present a spatio-temporal model that predicts image and video saliency through the phase spectrum of the Quaternion Fourier Transform but the model lacks the necessary high-level information for accurate saliency prediction.
While a recent learning-based approach~\cite{liu2016spatio} extends the image domain to the spatio-temporal domain by using LSTMs, such models are specialized for video data, rendering
them unable to simultaneously model image saliency.

\subsubsection{Domain Adaptation.}
We focus on domain specific learning, a form of domain adaptation which enables a learning system to process data from different domains by separating domain-invariant (shared) and domain-specific (private) parameters \cite{Chang_2019_CVPR}.
Domain Separation Networks (DSN) \cite{Bousmalis2016a}, for instance, are autoencoders with additional private encoders.
Instead of an autoencoder, Tsai~\etal~\cite{Tsai2017} introduce an adversarial loss that enforces shared and private encoders networks.
Xiao~\etal~\cite{Xiao2016} propose Domain Guided Dropout that results in different sub-networks for each domain, and Rozantev~\etal~\cite{Rozantsev2019} train entirely separate networks for each domain, coupled through a similarity loss.
In contrast to using separate networks, the AdaBN method~\cite{Li2016a} adjusts the batch-normalization (BN) parameters of a shared network based on samples from a given target domain.
The DSBN method~\cite{Chang_2019_CVPR} generalizes this idea by training a separate set of BN parameters for each domain.
In general, these existing methods result in a large proportion of domain-specific parameters.
In contrast, we propose domain-adaptation techniques that are aimed to bridge the domain gap of saliency datasets with a maximum proportion of shared parameters.

\section{Unified Image and Video Saliency Modeling}
\label{sec:unified}

\subsection{Domain-Shift Modeling}
\label{sec:domain}

\begin{figure}[ht]
{
\tiny
\sffamily
\def\svgwidth{\columnwidth}
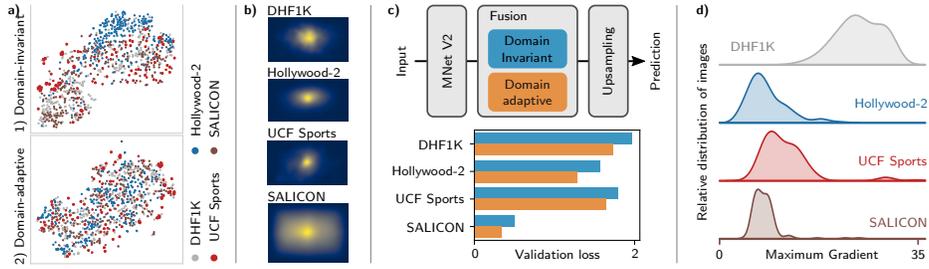
}
\caption{
Experiments to examine the domain shift between the saliency datasets.
\textbf{a)}~\mbox{t-SNE} visualization of MNet V2 features after domain-invariant and domain-adaptive normalization.
\textbf{b)}~Average ground truth saliency maps.
\textbf{c)}~Comparison of validation losses when training a simple saliency model with domain-invariant and domain-adaptive fusion.
\textbf{d)}~Distributions of ground truth saliency map sharpness.
}
\label{fig:prelim}
\end{figure}

In this section we present analyses to examine the domain shift between image and video data and between different video saliency datasets.
We use the insights to design corresponding domain adaptation methods.
Following Wang~\etal~\cite{wang2019revisiting}, we select the video saliency datasets DHF1K~\cite{wang2019revisiting}, Hollywood-2 and UCF Sports~\cite{Mathe2015}, and the image saliency dataset SALICON~\cite{jiang2015salicon}.

\subsubsection{Domain-Adaptive Batch Normalization.}
Batch normalization (BN) aims to reduce the internal covariate shift of neural network activations by transforming their distribution to zero mean and unit variance for each training batch.
Simultaneously, it computes running estimates of the distribution mean and variance for inference.
However, estimating these statistics across different domains results in inaccurate intra-domain statistics, and therefore a performance trade-off.
In order to examine the domain shift between the datasets, we conduct a simple experiment:
We randomly sample 256 images/frames from each dataset and compute their average pooled MobileNet V2 (MNet V2) features.
We then visualize the distribution of the feature vectors via t-SNE~\cite{maaten2008visualizing} after normalizing them with the mean and variance of 1) all samples (domain-invariant) or 2) the samples from the respective dataset (domain-adaptive).
The results, shown in Figure~\ref{fig:prelim}~a), reveal a significant domain shift among the different datasets, which is mitigated by the domain-adaptive normalization.
Consequently, we employ \emph{Domain-Adaptive Batch Normalization} (DABN), \ie, a different set of BN modules for each dataset.
During training and inference, each batch is constructed with data from one dataset and passed through the corresponding BN modules.

\subsubsection{Domain-Adaptive Priors.}
Figure~\ref{fig:prelim}~b) shows the average ground truth saliency map for each training dataset.
Among the video datasets, Hollywood-2 and UCF Sports exhibit the strongest center bias, which is plausible since they are biased towards certain content (movies and sports) while DHF1K is more diverse.
SALICON has a much weaker center bias than the video saliency datasets, which can potentially be explained by the longer viewing time of each image/frame (\SI{5}{\second} \vs \SIrange{30}{42}{\milli\second}) that allows secondary stimuli to be fixated.
Accordingly, we propose to learn a separate set of Gaussian prior maps for each dataset.

\subsubsection{Domain-Adaptive Fusion.}
\label{sec:dataAdapt}
We hypothesize that similar image features can have varying visual saliency for images/frames from different training datasets.
For example, the Hollywood-2 and UCF Sports datasets are \emph{task-driven}, \ie, the viewer is instructed to identify the main action shown.
On the other hand, the DHF1K and SALICON datasets contains \emph{free-viewing} fixations.
To test the hypothesis, we design a simple saliency predictor (see Figure~\ref{fig:prelim}~c):
The outputs of the MNet V2 model are fused to a single map by a \emph{Fusion} layer ($1\,{\times}\,1$ convolution) and upsampled through bilinear interpolation.
We train the \emph{Fusion} layer until convergence with 1) one set of weights (domain-invariant) or 2) different weights for each dataset (domain-adaptive).
We find that the validation loss is lower for all datasets for setting 2), where the network can weigh the importance of the feature maps differently for each dataset.
Consequently, we propose to learn a different set of \emph{Fusion} layer weights for each dataset.

\subsubsection{Domain-Adaptive Smoothing.}
\label{sec:dataSmoothing}
The size of the blurring filter which is used to generate the ground truth saliency maps from fixation maps can vary between datasets, especially since the images/frames are resized by different amounts.
To examine this effect, we compute the distribution of the ground truth saliency map sharpness for each dataset.
Sharpness is computed as the maximum image gradient magnitude after resizing to the model input resolution.
The results in Figure~\ref{fig:prelim}~d) confirm the heterogeneous distributions across datasets, revealing the highest sharpness for DHF1K.
Therefore, we propose to blur the network output with a different learned \emph{Smoothing} kernel for each dataset.

\begin{figure*}[t!]
{
\tiny
\sffamily
\centering
\includegraphics[width=\textwidth]{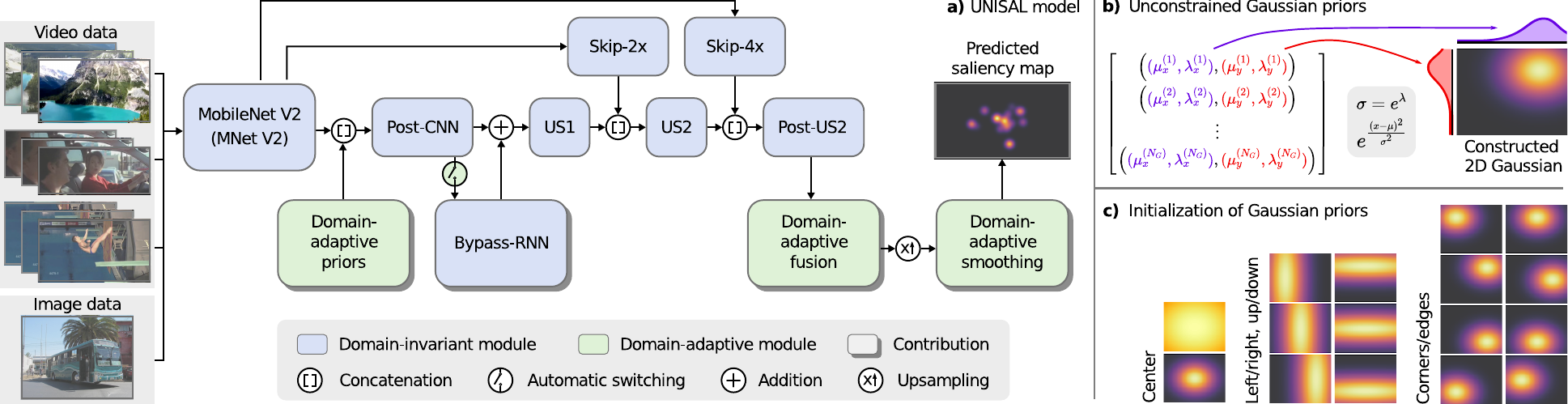}
}
\caption{
\textbf{a)} Overview of the proposed framework.
The model consists of a MobileNet V2 (MNet V2) encoder, followed by concatenation with learned Gaussian prior maps, a \emph{Bypass-RNN}, a decoder network with skip connections, and \emph{Fusion} and \emph{Smoothing} layers.
The prior maps, fusion, smoothing and batch-normalization modules are domain-adaptive in order to account for domain-shift between the image and video saliency datasets and enable high-quality shared features.
\textbf{b)} Construction of the prior maps from learned Gaussian parameters.
\textbf{c)} Prior maps initialization.
}
\label{fig:arch}
\end{figure*}

\subsection{UNISAL Network Architecture}
\label{sec:baseline}
We introduce a simple and lightweight neural network architecture termed \emph{UNISAL} that is designed to model image and video saliency coequally
and implements the proposed domain-adaptation techniques.
The architecture, illustrated in Figure~\ref{fig:arch}, follows an encoder-RNN-decoder design tailored for saliency modeling.

\subsubsection{Encoder Network.}
We use MobileNet-V2 (MNet V2)~\cite{sandler2018mobilenetv2} as our backbone encoder for three reasons: First, its small memory footprint enables training with sufficiently large sequence length and batch size; second, its small number of floating point operations allows for real time inference; and third, we expect the relatively small number of parameters to mitigate overfitting on smaller datasets like UCF Sports.
The main building blocks of MNet V2 are \emph{inverted residuals}, \ie, sequences of pointwise convolutions that decompress and compress the feature space, interleaved with depthwise separable $3{\times}3$ convolutions.
Overall, for an input resolution of $[r_x, r_y]$, MNet V2 computes feature maps at resolutions of $\frac{1}{2^\alpha}[r_x, r_y]$ with $\alpha\in\lbrace1,2,3,4,5\rbrace$.
The output has 1280 channels and scale $\alpha\,{=}\,5$.
Domain-Adaptive Batch Normalization is not used in MNet V2 since we initialize it with ImageNet-pretrained parameters.

\subsubsection{Gaussian Prior Maps.}
The domain-adaptive Gaussian prior maps are constructed at runtime from learned means and standard deviations.
The map with index $i=1,\dots,N_G$ is computed as
\begin{equation}\label{eq:gaussians}
g^{(i)}(x,y) = \gamma\,\mathrm{exp}\left(-\frac{(x - \mu_x^{(i)})^2}{(\sigma_x^{(i)})^2} -\frac{(y - \mu_y^{(i)})^2}{(\sigma_y^{(i)})^2}\right),
\end{equation}
where $\gamma=6$ is a scaling factor since the maps are concatenated with the ReLU6 activations of MNet V2.
In this formulation, if the standard deviation $\sigma^{(i)}_{xy}$ is optimized over $\mathbb{R}$, then the resulting variance $(\sigma^{(i)}_{xy})^2$ has the domain $\mathbb{R}_{\geq0}$, which can lead to division by zero.
Prior work which uses non-adaptive prior maps \cite{Cornia2016a} addresses this by clipping $\sigma^{(i)}_{xy}$ to a predefined interval $[a, b]$ with $a>0$ and clipping $\mu^{(i)}_{xy}$ to an interval around the center of the map.
However, these constraints potentially limit the ability to learn the optimal parameters.
Here, we propose \emph{unconstrained Gaussian prior maps} by substituting
$\sigma^{(i)}_{xy} = e^{\lambda^{(i)}_{xy}}$ and optimizing $\lambda^{(i)}_{xy}$ and $\mu^{(i)}_{xy}$ over $\mathbb{R}$.
Moreover, instead of drawing the initial Gaussian parameters from a normal distribution, which results in highly correlated maps, we initialize $N_G=16$ maps as shown in Figure~\ref{fig:arch}~c), covering a broad range of priors.
Finally, previous work usually introduces prior maps at the second to last layer in order to model the static center bias.
Here, we concatenate the prior maps with the encoder output before the RNN and decoder, in order to leverage the prior maps in higher-level features.

\subsubsection{Bypass-RNN.}
Modeling video saliency data requires a strategy to extract temporal features, such as an RNN, optical flow or 3D convolutions.
However, none of these techniques are generally suitable to process static inputs, whereas our goal is to process images and videos with one model.
Therefore, we introduce a \emph{Bypass-RNN}, \ie, a RNN whose output is added to its input features via a residual connection that is automatically omitted (bypassed) for static batches.
during training and inference.
Thus, the RNN only models the residual variations in visual saliency that are caused by
temporal features.

In the UNISAL model, the \emph{Bypass-RNN} is preceded by a \emph{post-CNN} module, which compresses the concatenated MNet V2 outputs and Gaussian prior maps to 256 channels.
For the Bypass-RNN, we use a convolutional GRU (\emph{cGRU}) RNN \cite{Valipour2016} due to its relative simplicity, followed by a pointwise convolution.
The cGRU has 256 hidden channels, $3{\times}3$ kernel size, recurrent dropout \cite{Gal2015} with probability $p=0.2$, and MobileNet-style convolutions, \ie, depthwise separable convolutions followed by pointwise convolutions.

\setlength{\tabcolsep}{6pt}
\begin{table}[t]
\centering
\caption{
Network modules and corresponding operations.
\emph{ConvDW}(\emph{c}) denotes a depthwise separable convolution with \emph{c} channels and kernel size $3{\times}3$, followed by batch normalization and ReLU6 activation.
\emph{ConvPW}($c_{\textrm{in}}$, $c_{\textrm{out}}$) is a pointwise $1{\times}1$ convolution with $c_\textrm{in}$ input and $c_\textrm{out}$ output channels, followed by batch normalization and, if $c_\textrm{in} \leq c_\textrm{out}$, by ReLU6 activation.
DO(\emph{p}) denotes 2D dropout with probability \emph{p}.
\emph{Up}(\emph{c}, \emph{n}) denotes \emph{n}-fold upsampling with bilinear interpolation of feature maps with \emph{c} channels.
}
\label{tab:modules}
\begin{tabular}{l  l}
\toprule
Module & Operations\\
\midrule
Post-CNN & \emph{ConvDW}(1280), \emph{ConvPW}(1280, 256)\\
Skip-4x & \emph{ConvPW}(64, 128), DO(0.6), \emph{ConvPW}(128, 64)\\
Skip-2x & \emph{ConvPW}(160, 256), DO(0.6), \emph{ConvPW}(256, 128)\\
US1 & \emph{Up}(256, 2)\\
US2 & \emph{ConvPW}(384, 768), \emph{ConvDW}(768), \emph{ConvPW}(768, 128), \emph{Up}(128, 2)\\
Post-US2 & \emph{ConvPW}(200, 400), \emph{ConvDW}(400), \emph{ConvPW}(400, 64)\\
Fusion & \emph{ConvPW(64, 1)}\\
\bottomrule
\end{tabular}
\end{table}
\setlength{\tabcolsep}{1.4pt}

\subsubsection{Decoder Network and Smoothing.}
The details of the decoder modules are listed in Table~\ref{tab:modules}.
First, the Bypass-RNN features are upsampled to scale $\alpha\,{=}\,4$ by \emph{US1} and concatenated with the output of \emph{Skip-2x}.
Next, the concatenated feature maps are upsampled to scale $\alpha\,{=}\,3$ by \emph{US2} and concatenated with the output of \emph{Skip-4x}.
The \emph{Post-US2} features are reduced to a single channel by an \emph{Domain-Adaptive Fusion} layer ($1\,{\times}\,1$ convolution) and upsampled to the input resolution via nearest-neighbor interpolation.
The upsampling is followed by a \emph{Domain-Adaptive Smoothing} layer with $41{\times}41$ convolutional kernels that explicitly models the dataset-dependent blurring of the ground-truth saliency maps.
Finally, following Jetley \etal \cite{Jetley2016}, we transform the output into a generalized Bernoulli distribution by applying a softmax operation across all output values.

\subsection{Domain-Aware Optimization}
\label{domain_optim}
\subsubsection{Domain-Adaptive Input Resolution.}
The images/frames have different aspect ratios for each dataset, specifically 4:3 for SALICON, 16:9 for DHF1K, 1.85:1 (median) for Hollywood-2, and 3:2 (median) for UCF Sports.
Our network architecture is fully-convolutional, and therefore agnostic to exact the input resolution.
Moreover, each mini-batch is constructed from one dataset due to DABN.
Therefore, we use input resolutions of $288{\times}384$, $224{\times}384$, $224{\times}416$ and $256{\times}384$ for SALICON, DHF1K, Hollywood-2 and UCF Sports, respectively.

\subsubsection{Assimilated Frame Rate.}
The frame rate of the DHF1K videos is \SI{30}{\fps} compared to \SI{24}{\fps} for Hollywood-2 and UCF Sports.
In order to assimilate the frame rates during training, and to train on longer time intervals, we construct clips using every \nth{5} frame for DHF1K and every \nth{4} frame for all others, yielding \SI{6}{\fps} overall.
During inference, the predictions are interleaved.

\section{Experiments}
\label{sec:experiments_main}
In this section, we compare the proposed method with current state-of-the-art image and video saliency models and provide detailed analyses are presented to gain an understanding of the proposed approach.

\subsection{Experimental Setup}
\label{sec:ExpSet}
\subsubsection{Datasets and Evaluation Metrics.}
To evaluate our proposed unified image and video saliency modeling framework, we jointly train UNISAL on datasets from both modalities.
For fair comparison, we use the same training data as ~\cite{Wang_2018_CVPR}, i.e., the SALICON~\cite{jiang2015salicon} image saliency dataset and the Hollywood-2~\cite{Mathe2015}, UCF Sports~\cite{Mathe2015}, and DHF1K~\cite{Wang_2018_CVPR} video saliency datasets.
For SALICON, we use the official training/validation/testing split of 10,000/5,000/5,000.
For Hollywood-2 and UCF Sports, we use the training and testing splits of 823/884 and 103/47 videos, and the corresponding validation sets are randomly sampled 10\% from the training sets, following~\cite{Wang_2018_CVPR}.
Hollywood-2 videos are divided into individual shots.
For DHF1K, we use the official training/validation/testing splits of 600/100/300 videos.
We compare against the state-of-the-art methods listed in~\cite{Wang_2018_CVPR} and add newer models with available implementations \cite{Min_2019_ICCV,lai2019video,Linardos2019,Cornia2016a,yang2019dilated}.
Moreover, test on the MIT300 benchmark \cite{Judd2012}, after fine-tuning with the MIT1003 dataset as suggested by the benchmark authors.
As in prior work~\cite{borji2012state,Wang_2018_CVPR}, we use the evaluation metrics AUC-Judd (AUC-J), Similarity Metric (SIM), shuffled AUC (s-AUC), Linear Correlation Coefficient (CC), and Normalized Scanpath Saliency (NSS) \cite{Bylinskii2019}.

\subsubsection{Implementation Details.}
We optimize the network via Stochastic Gradient Descent with momentum of 0.9 and weight decay of $10^{-4}$.
Gradients are clipped to $\pm2$.
The learning rate is set to 0.04 and exponentially decayed by a factor of 0.8 after each epoch.
The batch size is set to 4 for video data and 32 for SALICON.
The video clip length is set to 12 frames that are sampled as described in Section~\ref{domain_optim}.
Videos that are too short are discarded for training, which applies to Hollywood-2.
For comparability, we use the same loss formulation as Wang~\etal~\cite{wang2019revisiting}.
The model is trained for 16 epochs and with early stopping on the DHF1K validation set.
To prevent overfitting, the weights of MNet V2 are frozen for the first two epochs and afterwards trained with a learning rate that is reduced by a factor of 10.
The pretrained BN statistics of MNet V2 are frozen throughout training.
To account for dataset imbalance, the learning rate for SALICON batches is reduced by a factor of 2.
Our model is implemented using the PyTorch framework and trained on a NVIDIA GTX 1080 Ti GPU.
\begin{table*}[ht]
\caption{
Quantitative performance on the video saliency datasets.
The training settings (i) to (vi) denote training with:
(i) DHF1K, (ii) Hollywood-2, (iii) UCF Sports, (iv) SALICON, (v) DHF1K+Hollywood-2+UCF Sports, and (vi) DHF1K+Hollywood-2+UCF Sports+SALICON. Best performance is shown in \textbf{bold} while the second best is \underline{underlined}. The * symbol denotes training under setting (vi), while $\dagger$ indicates that the method is fine-tuned for each dataset.
}
\label{tab:bench}
\resizebox{\textwidth}{!}{%
\begin{tabular}{@{}l|l|ccccc|ccccc|ccccc@{}}
\toprule
\multirow{2}{*}{} & \multirow{2}{*}{\diagbox{Method}{Dataset}} & \multicolumn{5}{c|}{DHF1K} & \multicolumn{5}{c|}{Hollywood-2} & \multicolumn{5}{c}{UCF Sports} \\ \cmidrule(l){3-17}
 &  & AUC-J & SIM & s-AUC & CC & NSS & AUC-J & SIM & s-AUC & CC & NSS & AUC-J & SIM & s-AUC & CC & NSS \\ \midrule
\multirow{14}{*}{\rotatebox[origin=c]{90}{Dynamic models}} & PQFT~\cite{guo2009novel} & 0.699 & 0.139 & 0.562 & 0.137 & 0.749 & 0.723 & 0.201 & 0.621 & 0.153 & 0.755 & 0.825 & 0.250 & 0.722 & 0.338 & 1.780 \\
& Seo \etal~\cite{seo2009static} & 0.635 & 0.142 & 0.499 & 0.070 & 0.334 & 0.652 & 0.155 & 0.530 & 0.076 & 0.346 & 0.831 & 0.308 & 0.666 & 0.336 & 1.690 \\
& Rudoy \etal~\cite{rudoy2013learning} & 0.769 & 0.214 & 0.501 & 0.285 & 1.498 & 0.783 & 0.315 & 0.536 & 0.302 & 1.570 & 0.763 & 0.271 & 0.637 & 0.344 & 1.619 \\
& Hou \etal~\cite{hou2009dynamic} & 0.726 & 0.167 & 0.545 & 0.150 & 0.847 & 0.731 & 0.202 & 0.580 & 0.146 & 0.684 & 0.819 & 0.276 & 0.674 & 0.292 & 1.399 \\
& Fang \etal~\cite{fang2014video} & 0.819 & 0.198 & 0.537 & 0.273 & 1.539 & 0.859 & 0.272 & 0.659 & 0.358 & 1.667 & 0.845 & 0.307 & 0.674 & 0.395 & 1.787 \\
& OBDL~\cite{hossein2015many} & 0.638 & 0.171 & 0.500 & 0.117 & 0.495 & 0.640 & 0.170 & 0.541 & 0.106 & 0.462 & 0.759 & 0.193 & 0.634 & 0.234 & 1.382 \\
& AWS-D~\cite{leboran2016dynamic} & 0.703 & 0.157 & 0.513 & 0.174 & 0.940 & 0.694 & 0.175 & 0.637 & 0.146 & 0.742 & 0.823 & 0.228 & 0.750 & 0.306 & 1.631 \\
& OM-CNN~\cite{Jiang2018} & 0.856 & 0.256 & 0.583 & 0.344 & 1.911 & 0.887 & 0.356 & 0.693 & 0.446 & 2.313 & 0.870 & 0.321 & 0.691 & 0.405 & 2.089 \\
& Two-stream~\cite{bak2017spatio} & 0.834 & 0.197 & 0.581 & 0.325 & 1.632 & 0.863 & 0.276 & 0.710 & 0.382 & 1.748 & 0.832 & 0.264 & 0.685 & 0.343 & 1.753 \\
& *ACLNet~\cite{wang2019revisiting} & 0.890 & 0.315 & 0.601 & 0.434 & 2.354 & 0.913 & \underline{0.542} & 0.757 & 0.623 & 3.086 & 0.897 & 0.406 & 0.744 & 0.510 & 2.567 \\
& TASED-Net~\cite{Min_2019_ICCV} & 0.895 & 0.361 & 0.712 & 0.470 & 2.667 & 0.918 & 0.507 & 0.768 & 0.646 & 3.302 & 0.899 & 0.469 & 0.752 & 0.582 & 2.920 \\
& STRA-Net~\cite{lai2019video} & 0.895 & 0.355 & 0.663 & 0.458 & 2.558 & 0.923 & 0.536 & \underline{0.774} & 0.662 & 3.478 & {0.910} & 0.479 & 0.751 & 0.593 & 3.018 \\
& $\dagger$SalEMA~\cite{Linardos2019} & 0.890 & \textbf{0.465} & 0.667 & 0.449 & 2.573 & 0.919 & 0.487 & 0.708 & 0.613 & 3.186 & 0.906 & 0.431 & 0.740 & 0.544 & 2.638 \\
& *SalEMA~\cite{Linardos2019} & 0.895 & {0.283} & \textbf{0.739} & 0.414 & 2.285 & 0.875 & 0.371 & 0.663 & 0.456 & 2.214 & 0.899 & 0.381 & 0.769 & 0.521 & 2.503 \\
\midrule
\multirow{9}{*}{\rotatebox[origin=c]{90}{Static models}} & ITTI~\cite{itti1998model} & 0.774 & 0.162 & 0.553 & 0.233 & 1.207 & 0.788 & 0.221 & 0.607 & 0.257 & 1.076 & 0.847 & 0.251 & 0.725 & 0.356 & 1.640 \\
 & GBVS~\cite{harel2007graph} & 0.828 & 0.186 & 0.554 & 0.283 & 1.474 & 0.837 & 0.257 & 0.633 & 0.308 & 1.336 & 0.859 & 0.274 & 0.697 & 0.396 & 1.818 \\
 & SALICON~\cite{huang2015salicon} & 0.857 & 0.232 & 0.590 & 0.327 & 1.901 & 0.856 & 0.321 & 0.711 & 0.425 & 2.013 & 0.848 & 0.304 & 0.738 & 0.375 & 1.838 \\
 & Shallow-Net~\cite{pan2016shallow} & 0.833 & 0.182 & 0.529 & 0.295 & 1.509 & 0.851 & 0.276 & 0.694 & 0.423 & 1.680 & 0.846 & 0.276 & 0.691 & 0.382 & 1.789 \\
 & Deep-Net~\cite{pan2016shallow} & 0.855 & 0.201 & 0.592 & 0.331 & 1.775 & 0.884 & 0.300 & 0.736 & 0.451 & 2.066 & 0.861 & 0.282 & 0.719 & 0.414 & 1.903 \\
 & *Deep-Net~\cite{pan2016shallow} & 0.874 & 0.288 & 0.610 & 0.374 & 1.983 & 0.901 & 0.482 & 0.740 & 0.597 & 2.834 & 0.880 & 0.365 & 0.729 & 0.475 & 2.448 \\
 & DVA~\cite{wang2017deep} & 0.860 & 0.262 & 0.595 & 0.358 & 2.013 & 0.886 & 0.372 & 0.727 & 0.482 & 2.459 & 0.872 & 0.339 & 0.725 & 0.439 & 2.311 \\
 & *DVA~\cite{wang2017deep} & 0.883 & 0.297 & 0.623 & 0.397 & 2.237 & 0.907 & 0.497 & 0.753 & 0.607 & 2.942 & 0.892 & 0.387 & 0.740 & 0.492 & 2.503 \\
 & SalGAN~\cite{pan2017salgan} & 0.866 & 0.262 & {0.709} & 0.370 & 2.043 & 0.901 & 0.393 & \textbf{0.789} & 0.535 & 2.542 & 0.876 & 0.332 & 0.762 & 0.470 & 2.238 \\ \midrule
\multirow{6}{*}{\rotatebox[origin=c]{90}{UNISAL (ours)}} & Training setting (i) & \underline{0.899} & 0.378 & 0.686 & 0.481 & 2.707 & 0.920 & 0.496 & 0.710 & 0.612 & 3.279 & 0.896 & 0.443 & 0.717 & 0.553 & 2.689 \\
& Training setting (ii) & 0.881 & 0.313 & 0.690 & 0.422 & 2.352 & \underline{0.932} & 0.534 & 0.762 & 0.672 & {3.803} & 0.892 & 0.440 & 0.735 & 0.566 & 2.768 \\
& Training setting (iii) & 0.869 & 0.286 & 0.664 & 0.375 & 2.056 & 0.890 & 0.392 & 0.683 & 0.475 & 2.350 & 0.908 & 0.502 & 0.764 & 0.614 & {3.076} \\
& Training setting (iv) & 0.883 & 0.288 & \underline{0.715} & 0.410 & 2.259 & 0.912 & 0.432 & 0.750 & 0.565 & 2.897 & 0.892 & 0.428 & \underline{0.776} & 0.561 & 2.740 \\
& Training setting (v) & \textbf{0.901} & {0.384} & 0.692 & \underline{0.488} & \underline{2.739} & \textbf{0.934} & \textbf{0.544} & 0.758 & \textbf{0.675} & \textbf{3.909} & \underline{0.917} & \underline{0.514} & \textbf{0.786} & \underline{0.642} & \underline{3.260} \\
& Training setting (vi) & \textbf{0.901} & \underline{0.390} & 0.691 & \textbf{0.490} & \textbf{2.776} & \textbf{0.934} & \underline{0.542} & {0.759} & \underline{0.673} & \underline{3.901} & \textbf{0.918} & \textbf{0.523} & {0.775} & \textbf{0.644} & \textbf{3.381} \\ \bottomrule
\end{tabular}%
}
\end{table*}

\subsection{Quantitative Evaluation}
\label{sec:quantitative}
The results of the quantitative evaluation are shown in Table~\ref{tab:bench} for the video saliency datasets and in Tables~\ref{tab:benchS} and \ref{tab:compS} for the image datasets.
For video saliency prediction, in order to analyze the impact of---and generalization across---different datasets, we evaluate six training settings:
i) DHF1K, ii) Hollywood-2, iii) UCF Sports, iv) SALICON, v) DHF1K, Hollywood-2, and UCF Sports, vi) DHF1K, Hollywood-2, UCF Sports and SALICON.
For fair comparison, we include state-of-the-art methods that are trained on our best-performing training setting (iv):
The ACLNet~\cite{wang2019revisiting} video saliency model and the Deep-Net~\cite{pan2016shallow} and DVA~\cite{wang2017deep} image saliency models.
In addition, we provide the performance of SalEMA~\cite{Linardos2019}, which is based on SalGAN~\cite{pan2017salgan}, after fine-tuning the model with training setting (vi).
Other state-of-the-art video saliency models \cite{Jiang2018,Min_2019_ICCV,lai2019video} are not suitable for training with image data as discussed in Section~\ref{sec:introduction}.
We observe that the proposed UNISAL model significantly outperforms previous static and dynamic methods, across almost all metrics.
We obtain the following additional findings:
1) Training with all video saliency datasets (setting (v)) \emph{always} improves performance compared to individual video saliency datasets (settings (i) to (iii)).
This has not been the case for UCF Sports in a previous cross-dataset evaluation study \cite{wang2019revisiting}.
2) Additionally including image saliency data (setting (vi)) further improves performance for most metrics for DHF1K and \mbox{UCF Sports}.
The exception is Hollywood-2, but the performance decrease is less than 1\%.

For image saliency prediction, UNISAL performs on par with state-of-the-art image saliency models both on the SALICON and MIT300 benchmark as shown in Table~\ref{tab:benchS}.
In addition, we evaluate state-of-the-art video saliency models on SALICON dataset as shown in Table \ref{tab:compS}.
For ACLNet~\cite{wang2019revisiting} we use the auxiliary output which is trained on SALICON (using the LSTM output yielded worse performance).
For SalEMA~\cite{Linardos2019}, we fine-tuned their best performing model with training setting (vi).
A large performance jump can be observed for the domain-adaptive UNISAL model.

\begin{figure*}[t]
\centering
\includegraphics[width=\textwidth]{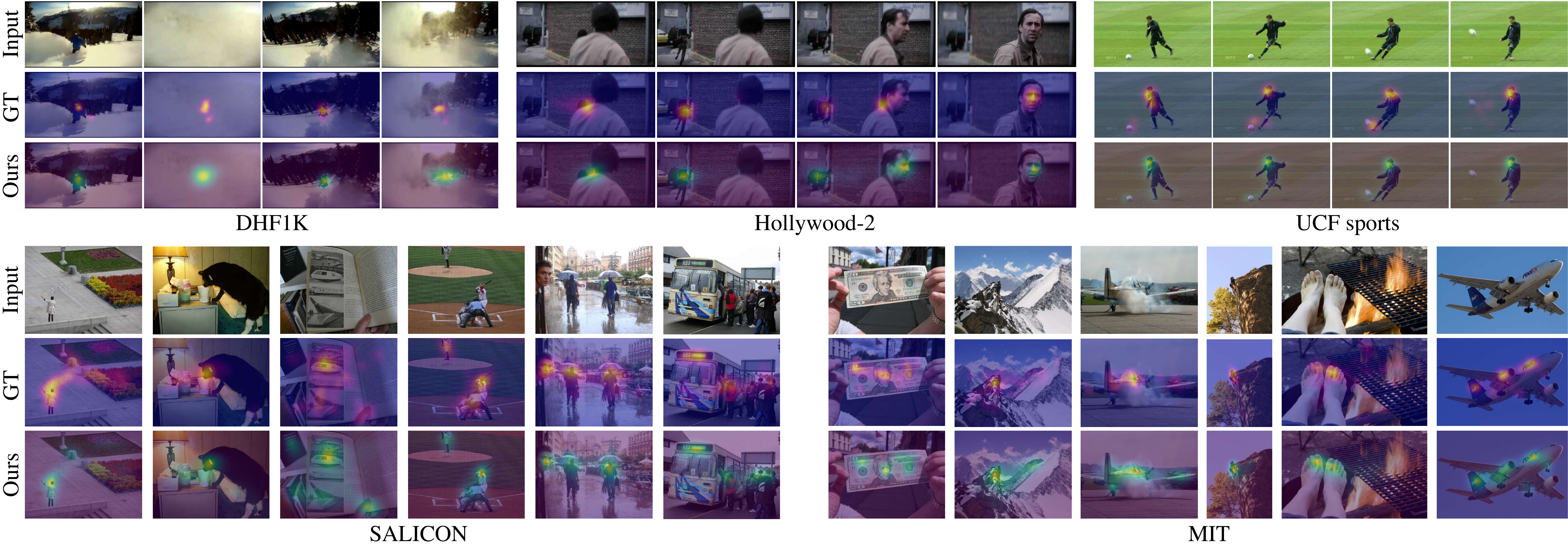}
\caption{Qualitative performance of the proposed approach on video (top part) and image (bottom part) saliency prediction.}
\label{fig:quali}
\end{figure*}

\begin{table}[t]
\begin{minipage}[t]{.57\linewidth}
\caption{
performance on the SALICON and MIT300 benchmarks. Best performance is shown in \textbf{bold} while the second best is \underline{underlined}. Training setting (vi) is used for UNISAL. See supplementary material for other settings and updated MIT300 results.
}
\label{tab:benchS}
\resizebox{\linewidth}{!}{%
\begin{tabular}{@{}l|ccccc|ccccc@{}}
\toprule
\multirow{2}{*}{\diagbox{Method}{Dataset}} & \multicolumn{5}{c|}{SALICON} & \multicolumn{5}{c}{MIT300} \\
\cmidrule(l){2-11}
& AUC-J & SIM & s-AUC & CC & NSS & AUC-J & SIM & s-AUC & CC & NSS \\ \midrule
ITTI~\cite{itti1998model} & 0.667 & 0.378 & 0.610 & 0.205 & - & 0.75 & 0.44 & 0.63 & 0.37 & 0.97 \\
GBVS~\cite{harel2007graph} & 0.790 & 0.446 & 0.630 & 0.421 & - & 0.81 & 0.48 & 0.63 & 0.48 & 1.24 \\
SALICON~\cite{huang2015salicon} & - & - & - & - & - & \textbf{0.87} & 0.60 & \textbf{0.74} & \underline{0.74} & \underline{2.12} \\
Shallow-Net~\cite{pan2016shallow} & {0.836} & \underline{0.520} & 0.670 & 0.596 & - & 0.80 & 0.46 & 0.64 & 0.53 & - \\
Deep-Net~\cite{pan2016shallow} & - & - & 0.724 & 0.609 & 1.859 & 0.83 & 0.52 & 0.69 & 0.58 & 1.51 \\
SAM-ResNet~\cite{Cornia2016a} & \textbf{0.883} & - & \underline{0.779} & 0.842 & \underline{3.204} & \textbf{0.87} & \textbf{0.68} & 0.70 & \textbf{0.78} & \textbf{2.34} \\
DVA~\cite{wang2017deep} & - & - & - & - & - & 0.85 & 0.58 & 0.71 & 0.68 & 1.98 \\
DINet~\cite{yang2019dilated} & \textbf{0.884} & - & \textbf{0.782} & \underline{0.860} & \textbf{3.249} & \underline{0.86} & - & 0.71 & \textbf{0.79} & \textbf{2.33} \\
SalGAN~\cite{pan2017salgan} & - & - & {0.772} & {0.781} & {2.459} & \underline{0.86} & \underline{0.63} & \underline{0.72} & 0.73 & 2.04 \\
\midrule
UNISAL (ours) & \underline{0.864} & \textbf{0.775} & {0.739} & \textbf{0.879} & {1.952} & \textbf{0.872} & \textbf{0.674} & \textbf{0.743} & \textbf{0.784} & \textbf{2.322} \\
\bottomrule
\end{tabular}%
}
\end{minipage}
\hfill
\begin{minipage}[t]{.38\linewidth}
\caption{
Comparison for dynamic models on the static SALICON benchmark. Best performance is shown in \textbf{bold} while the second best is \underline{underlined}. Training setting (vi) is used for all methods.
}
\label{tab:compS}
\resizebox{\linewidth}{!}{%
\begin{tabular}{@{}l|ccccc@{}}
\toprule
Method & AUC-J & SIM & s-AUC & CC & NSS \\ \midrule
SalEMA~\cite{Linardos2019} & 0.732 & 0.470 & 0.519 & 0.411 & 0.760 \\
ACLNet~\cite{wang2019revisiting} & 0.843 & 0.688 & \underline{0.698} & 0.771 & 1.618 \\
UNISAL (w/o DA) & \underline{0.848} & \underline{0.690} & 0.676 & \underline{0.799} & \underline{1.654} \\
UNISAL (final) & \textbf{0.864} & \textbf{0.775} & \textbf{0.739} & \textbf{0.879} & \textbf{1.952} \\
\bottomrule
\end{tabular}
}
\end{minipage}
\end{table}

\subsection{Qualitative Evaluation}
\label{sec:qualitative}
In Figure~\ref{fig:quali}, we show randomly selected saliency predictions for both images and videos.
It is visible that the proposed unified model performs well on both modalities.
For challenging dynamic scenes with complete occlusion (DHF1K, left), the model correctly memorizes the salient object location, indicating that long-term temporal dependencies are effectively modeled.
Moreover, the model correctly predicts shifting observer focus in the presence of multiple salient objects, as evident from the \mbox{Hollywood-2} and UCF Sports samples.
The results on static scenes (bottom part of Figure~\ref{fig:quali}) confirm that the proposed unified model indeed generalizes to static scenes.


\begin{table}[t]
\centering
\caption{
Ablation study of the proposed approach on the DHF1K and SALICON validation sets.
The proposed components are added incrementally to the baseline to quantify their contribution.
Training setting (vi) is used for this study.
}
\label{tab:ablation}
\begin{adjustbox}{max width=1.0\textwidth}
\begin{tabular}{@{}l|cccccc|cccccc@{}}
\toprule
\multirow{2}{*}{\diagbox{Config.}{Dataset}} & \multicolumn{6}{c|}{DHF1K} & \multicolumn{5}{c}{SALICON} \\
\cmidrule(l){2-13}
& KLD $\downarrow$ & AUC-J $\uparrow$ & SIM $\uparrow$ & s-AUC $\uparrow$ & CC $\uparrow$ & NSS $\uparrow$ & KLD $\downarrow$ & AUC-J $\uparrow$ & SIM $\uparrow$ & s-AUC $\uparrow$ & CC $\uparrow$ & NSS $\uparrow$ \\ \midrule
Baseline & 1.877 & 0.863 & 0.282 & 0.659 & 0.372 & 2.057 & 0.551 & 0.824 & 0.607 & 0.633 & 0.711 & 1.415 \\
+ Gaussian & 1.776 & 0.879 & 0.300 & 0.668 & 0.411 & 2.273 & 0.394 & 0.848 & 0.675 & 0.685 & 0.801 & 1.634 \\
+ RNNRes & 1.754 & 0.881 & 0.302 & 0.666 & 0.411 & 2.274 & 0.450 & 0.843 & 0.648 & 0.665 & 0.770 & 1.531 \\
+ SkipConnect & 1.749 & 0.884 & 0.308 & 0.658 & 0.412 & 2.301 & 0.404 & 0.841 & 0.673 & 0.664 & 0.777 & 1.600 \\
+ Smoothing & 1.770 & 0.882 & 0.295 & 0.677 & 0.416 & 2.305 & 0.369 & 0.848 & 0.690 & 0.676 & 0.799 & 1.654 \\
+ DomainAdaptive & 1.526 & 0.907 & 0.373 & 0.685 & 0.482 & 2.731 & 0.231 & 0.867 & 0.768 & 0.712 & 0.877 & 1.925 \\
Final & 1.531 & 0.907 & 0.381 & 0.691 & 0.487 & 2.755 & 0.226 & 0.867 & 0.771 & 0.725 & 0.880 & 1.923 \\ \bottomrule
\end{tabular}%
\end{adjustbox}
\end{table}

\subsection{Ablation Study}
\label{sec:ablation}
We analyze the contribution of each proposed component:
1) Gaussian prior maps; 2) RNN residual connection; 3) skip connections; 4) \emph{Smoothing} layer; 5) domain-adaptive operations (incl. \mbox{Bypass-RNN}); and 6) domain-aware optimization.
We perform the ablation on the representative DHF1K and SALICON validation sets.
The results in Table~\ref{tab:ablation} show that each of the proposed components contributes a considerable performance increase.
Overall, the domain-adaptive operations contribute the most, both for DHF1K and SALICON.
This indicates that mitigating the domain shift between datasets is a crucial component of UNISAL, confirming our initial studies in Section~\ref{sec:domain}.
The Gaussian prior maps yield the second largest gain, indicating the effectiveness of their proposed unconstrained optimization and early position in the model.


\begin{figure}[t]
\centering
\includegraphics[width=\columnwidth]{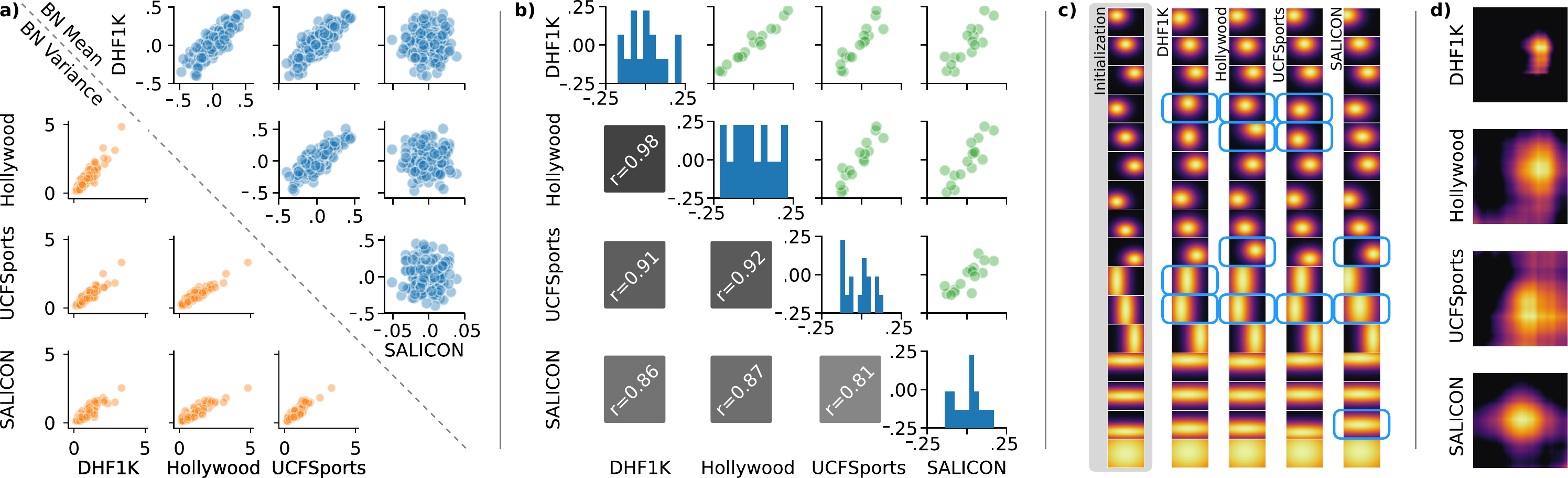}
\caption{
Retrospective analysis of the domain-adaptive modules.
a)~Correlation of the batch normalization statistics between datasets (\emph{US2} module, representative).
The upper-right plots correlate the estimated means and the lower-left plots the estimated variances.
b)~Correlation of the \emph{Fusion} layer weights between datasets.
The plots on the diagonal show the distribution of weights of the respective dataset.
The lower-left part shows Pearson's correlation coefficients.
c) Gaussian prior maps.
Significant deviations from the initialization are highlighted.
d)~\emph{Smoothing} kernel of each dataset.
}
\label{fig:sources}
\end{figure}

\subsection{Inter-Dataset Domain Shift}
Figure \ref{fig:sources} shows the retrospective analysis of the four domain-adaptive modules.
The DABN estimated means in Figure \ref{fig:sources}~a) are correlated among video datasets with Pearson correlation coefficients $r$ between 82\% to 83\%, but not correlated between SALICON and the video datasets ($r<3\%$).
Similarly, the DABN variances are least correlated between SALICON and the video datasets (90\% vs 92\%).
This confirms the shift of the feature distributions between datasets, especially between SALICON and the video data.
The domain-adaptive \emph{Fusion} layer weights shown in Figure~\ref{fig:sources} b) are generally correlated across datasets, with $r>81\%$.
However, as for the DABN, SALICON is the least correlated with the other datasets.
Moreover, many of the SALICON \emph{Fusion} weights lie near zero compared to the video datasets, which indicates that only a subset of the video saliency features is relevant for image saliency.
The \emph{Domain-Adaptive Fusion} layer models these differences while the remaining network weights are shared.
The domain-adaptive Gaussian prior maps shown in Figure~\ref{fig:sources} c) are successfully learned with our proposed unconstrained parametrization, as observed by the deviations from the initialization.
Some prior maps are similar across datasets while others vary visibly, indicating that the different domains have different optimal priors.
Finally, the learned \emph{Smoothing} kernels shown in Figure~\ref{fig:sources}~d) vary significantly across datasets.
As expected, the DHF1K dataset, which has the least blurry training targets, results in the most narrow \emph{Smoothing} filter.

\newcolumntype{R}{>{\raggedleft\arraybackslash}X}
\newcolumntype{L}{>{\raggedright\arraybackslash}X}

\begin{table}[t]
\centering
\caption{
Model size and runtime comparison of saliency prediction methods (based on the DHF1K benchmark \cite{wang2019revisiting}).
Best performance is shown in \textbf{bold}.
}
\label{tab:size_time}
\begin{tabular}{l r|l r}
\toprule
Method & Model size (MB) & Method & Runtime (s)\\
\midrule
Shallow-Net~\cite{pan2016shallow} & 2,500 & Two-stream~\cite{bak2017spatio} & 20\\
{STRA-Net~\cite{lai2019video}} & {641} & SALICON~\cite{huang2015salicon} & 0.5 \\
{SalEMA~\cite{Linardos2019}} & {364} &Shallow-Net~\cite{pan2016shallow} & 0.1\\
Two-stream~\cite{bak2017spatio} & 315 & DVA~\cite{wang2017deep} & 0.1\\
ACLNet~\cite{wang2019revisiting} & 250 & Deep-Net~\cite{pan2016shallow} & 0.08\\
SalGAN~\cite{pan2017salgan} & 130 & TASED-Net~\cite{Min_2019_ICCV} & 0.06\\
SALICON~\cite{huang2015salicon} & 117 & ACLNet~\cite{wang2019revisiting} & 0.02\\
Deep-Net~\cite{pan2016shallow} & 103 & SalGAN~\cite{pan2017salgan} & 0.02\\
DVA~\cite{wang2017deep} & 96 & STRA-Net~\cite{lai2019video} & 0.02\\
TASED-Net~\cite{Min_2019_ICCV} & 82 & SalEMA~\cite{Linardos2019} & 0.01\\
\midrule
UNISAL (ours) & \textbf{15.5} & UNISAL (ours) & \textbf{0.009}\\
\bottomrule
\end{tabular}
\end{table}

\subsection{Computational Load}
\label{sec:analysis}
With the design of ever more complex network architectures,
few studies evaluate the model size, although performance gains can often be traced back to more parameters.
We compare the size of UNISAL to the state-of-the-art video saliency predictors in the left column of Table~\ref{tab:size_time}.
Our model is the most light-weight by a significant margin, with over $5\times$ smaller size than TASED-Net, which is the current state-of-the-art on the DHF1K benchmark (see also Figure~\ref{fig:teaser}).
The same result applies when comparing to the deep image saliency methods from Table~\ref{tab:benchS}, whose sizes range from \SI{92}{\mega\byte} for DVA to \SI{2.5}{\giga\byte} for Shallow-Net.

Another key issue for real-world applications is the model efficiency.
Consequently, we present a GPU runtime comparison (processing time per frame) of video saliency models in the right column of Table~\ref{tab:size_time}.
Our model is the most efficient compared to previous state-of-the-art methods.
In addition, we observe a CPU (Intel Xeon W-2123 at 3.60GHz) runtime of \SI{0.43}{\second} (\SI{2.3}{\fps}), which is faster than some models' GPU runtime.
Considering both the model size and the runtime, the proposed saliency modeling approach achieves state-of-the-art performance in terms of real-world applicability.
While the MNet V2 encoder makes a large contribution to low model size and runtime, other contributing factors are: Separable convolutions throughout the cGRU and decoder; cGRU at the low-resolution bottleneck; bilinear upsampling.
Without these measures the model size and runtime increase to \SI{59.4}{\mega\byte} and \SI{0.017}{\second}, respectively.

\section{Discussion and Conclusion}
\label{sec:discussion}
In this paper, we have presented a simple yet effective approach to unify static and dynamic saliency modeling.
To bridge the domain gap, we found it crucial to account for different sources of inter-dataset domain shift through corresponding novel domain-adaptive modules.
We integrated the domain-adaptive modules into the new, lightweight and simple UNISAL architecture which is designed to model both data modalities coequally.
We observed state-of-the-art performance on video saliency datasets, and competitive performance on image saliency datasets, with a 5 to 20-fold reduction in model size compared to the \emph{smallest} previous deep model, and faster runtime.
We found that the domain-adaptive modules capture the differences between image and video saliency data, resulting in improved performance on each individual dataset through joint training.
We presented preliminary and retrospective experiments which explain the merit of the domain-adaptive modules.
To our knowledge, this is the first attempt towards unifying image and video saliency modeling in 
a single framework.
We believe that our work can serve as a basis for further research into joint modeling of these modalities.

\subsubsection{Acknowledgements.}
We acknowledge the EPSRC (Project Seebibyte, reference EP/M013774/1) and the NVIDIA Corporation for the donation of GPU.

\end{document}


\pagestyle{headings}
\mainmatter
\def\ECCVSubNumber{3601}  

\title{Unified Image and Video Saliency Modeling (Supplementary Material)}

%
\author{
Richard Droste\thanks{Richard Droste and Jianbo Jiao contributed equally to this work.} \and
Jianbo Jiao\textsuperscript{\thefootnote} \and
J.\ Alison Noble
}
%
\authorrunning{R. Droste et al.}
%
\institute{
University of Oxford\\
\email{\{richard.droste, jianbo.jiao, alison.noble\}@eng.ox.ac.uk}
}
\maketitle

\section{Introduction}
In this supplementary material, we provide additional quantitative and qualitative results for a better understanding of the proposed model for unified image and video saliency analysis.
The contents are structured as follows:
%
\begin{itemize}
	\item[] Section \ref{Video}: Additional Qualitative Video Saliency Results
	\item[] Section \ref{Image}: Additional Qualitative Image Saliency Results
	\item[] Section \ref{Cross}: Cross-Domain Predictions
	\item[] Section \ref{Bias}: Additional Center Bias Analysis
	\item[] Section \ref{Abla}: Additional Ablation Studies
	\item[] Section \ref{SALICON}: SALICON Cross-Dataset Generalization
  \item[] Section \ref{MIT300Prob}: MIT300 Probabilistic Benchmark Results
	\item[] Section \ref{Scoring}: Details for Quantitative Evaluation
	\item[] Section \ref{Code}: Code
\end{itemize}

\section{Additional Qualitative Video Saliency Results}\label{Video}
\begin{figure*}[t]
\centering
\includegraphics[width=\textwidth]{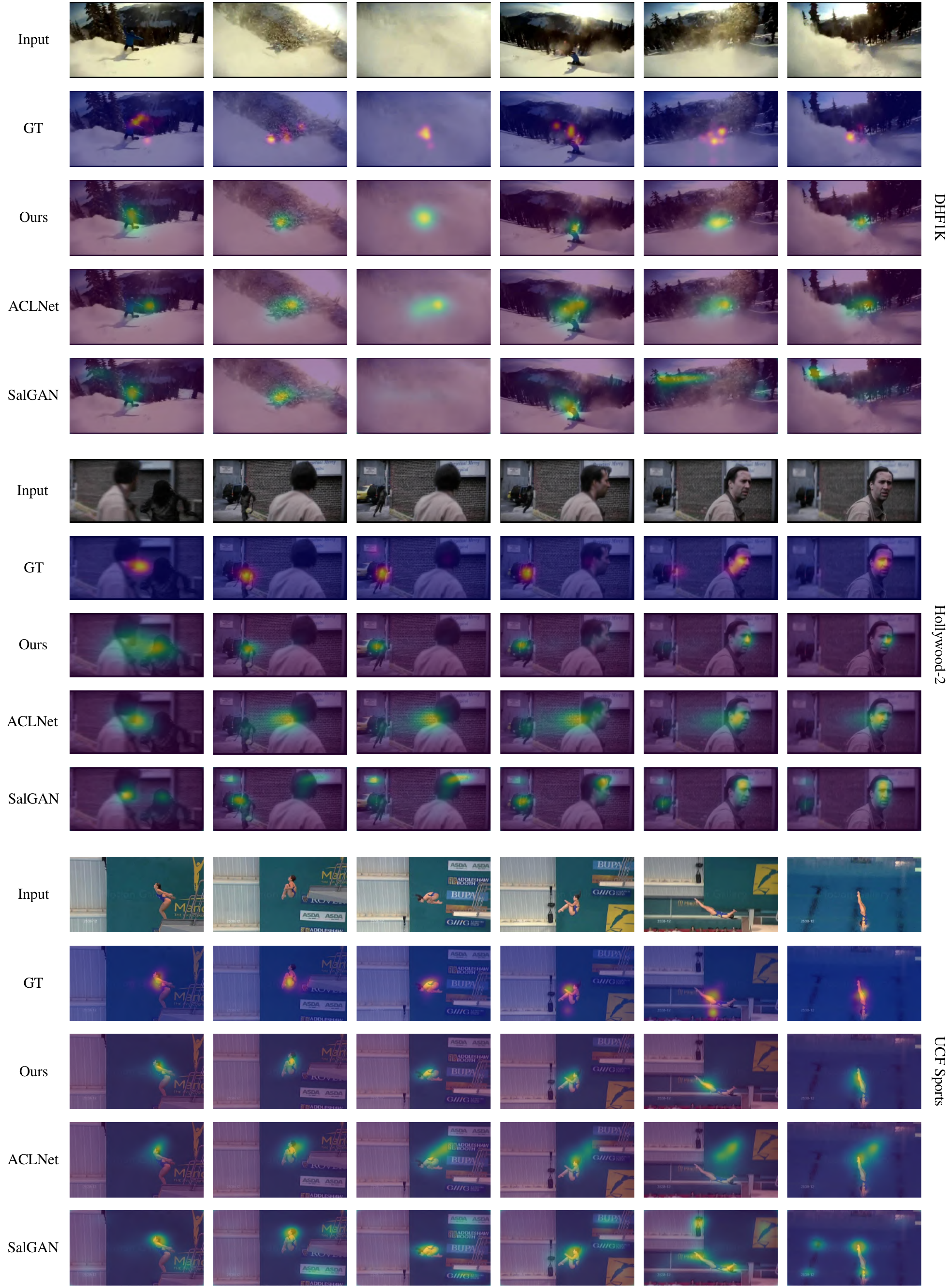}
\caption{
Additional qualitative video saliency prediction results.
Predictions of the proposed UNISAL model are compared to those of ACLNet~\cite{wang2019revisiting} and SalGAN~\cite{pan2017salgan}.
}
\label{fig:quali_vid}
\end{figure*}

We present further qualitative video saliency prediction results in addition to those shown in the main paper.
Also, we include comparisons to predictions generated with state-of-the-art methods~\cite{wang2019revisiting,pan2017salgan}.
Representative clips are sampled from the three video saliency datasets (DHF1K~\cite{wang2019revisiting}, UCF Sports~\cite{Mathe2015}, and Hollywood-2~\cite{Mathe2015}).
The results are shown in the supplementary video file \emph{3601-supp.mp4} (also available at \url{https://www.youtube.com/watch?v=4CqMPDI6BqE}).
Video frame-based examples are shown in Figure~\ref{fig:quali_vid}.

\section{Additional Qualitative Images Saliency Results}\label{Image}
\begin{figure*}[t]
\centering
\includegraphics[height=\textheight-1.5cm]{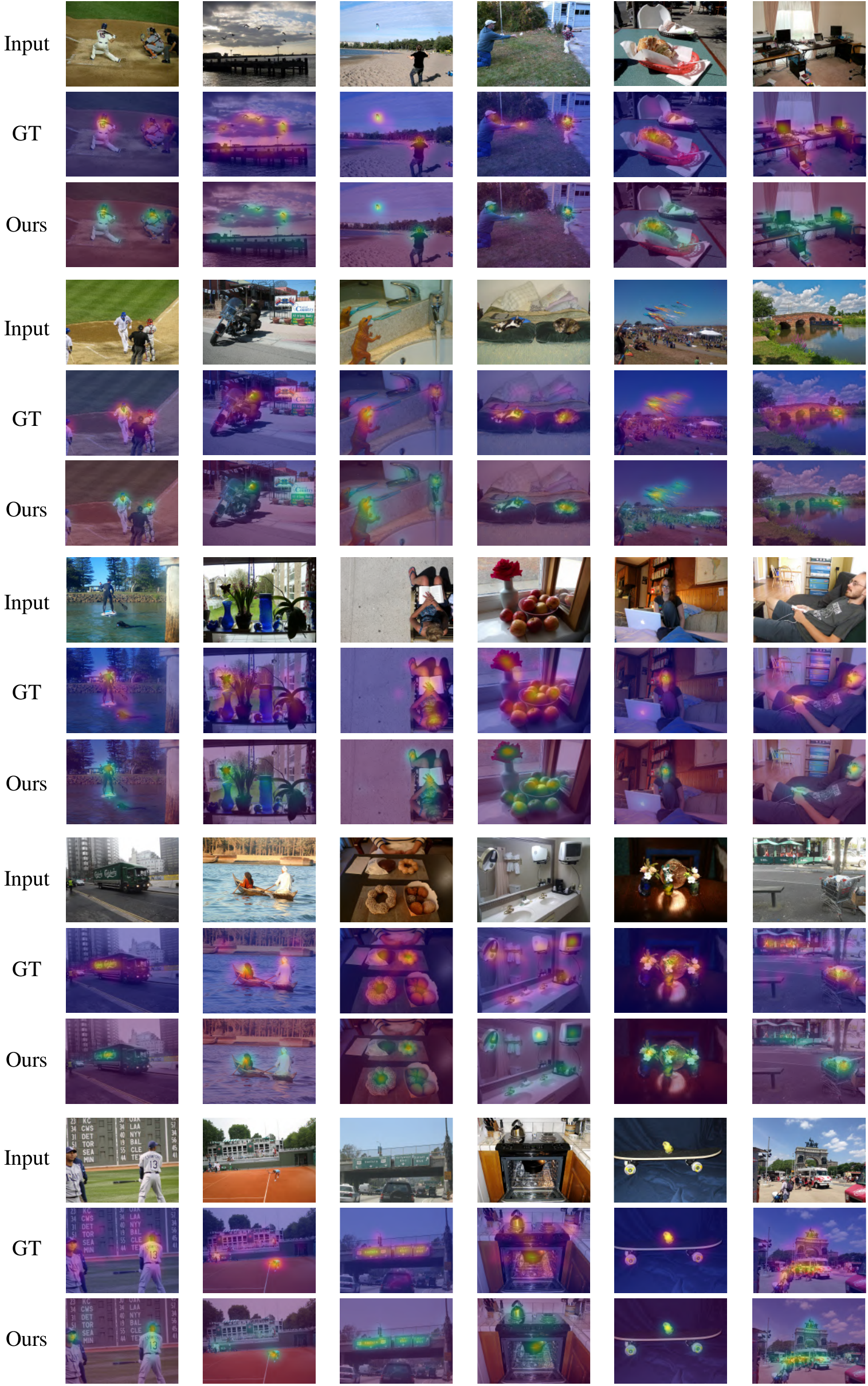}
\caption{
Additional qualitative image saliency prediction results of the proposed UNISAL model for the SALICON dataset.
}
\label{fig:quali_img}
\end{figure*}

\begin{figure*}[t]
\centering
\includegraphics[width=\textwidth]{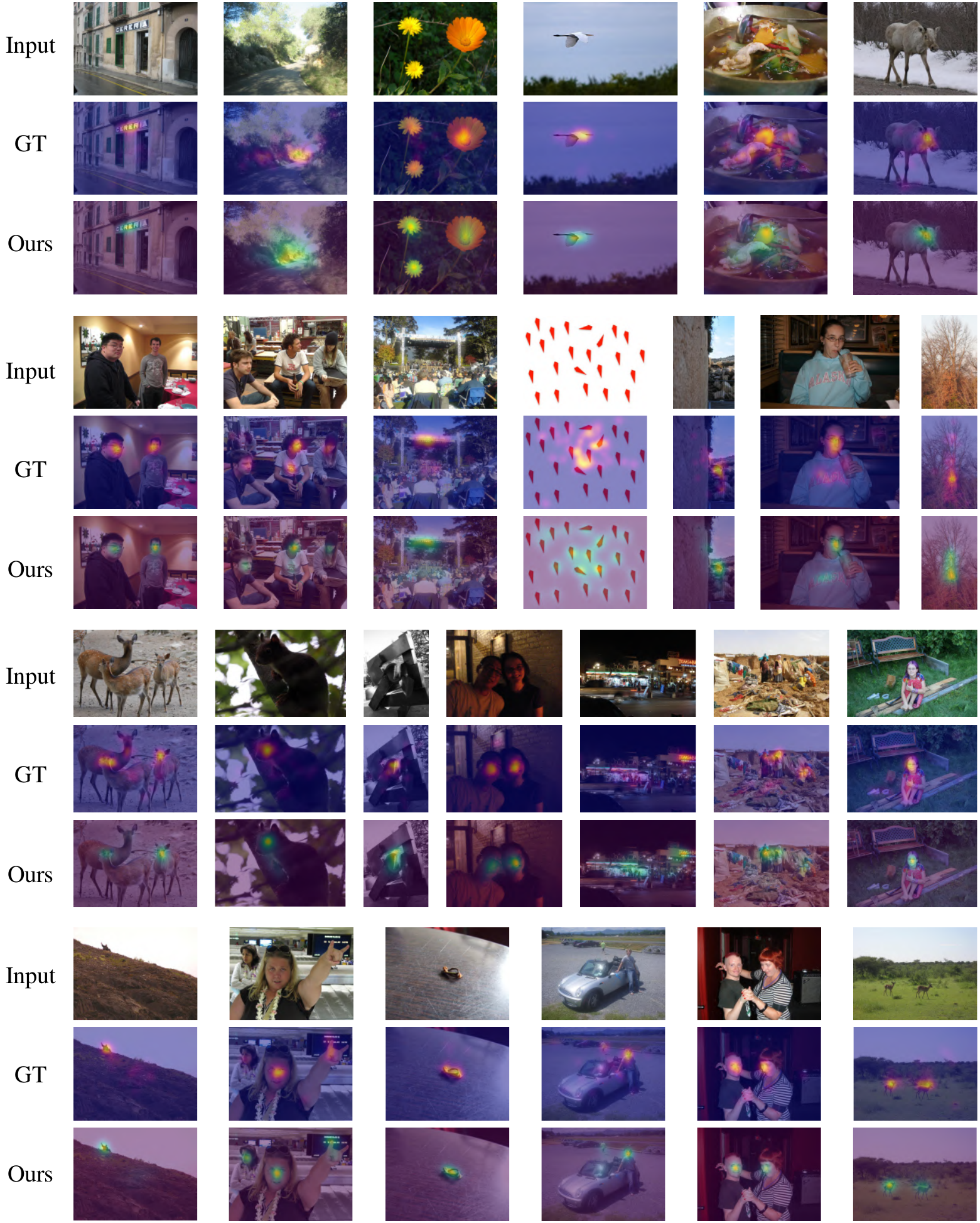}
\caption{
Additional qualitative image saliency prediction results of the proposed UNISAL model for the MIT1003 dataset.
}
\label{fig:quali_img2}
\end{figure*}

We include further qualitative image saliency prediction results in addition to those presented in the main paper.
Representative images are sampled from the SALICON~\cite{jiang2015salicon} and MIT1003~\cite{Judd2012} datasets.
The results are shown in Figure~\ref{fig:quali_img} and Figure~\ref{fig:quali_img2} for SALICON and MIT1003, respectively.

\section{Cross-Domain Predictions}
\label{Cross}
\begin{figure*}[t]
\centering
\includegraphics[width=\textwidth]{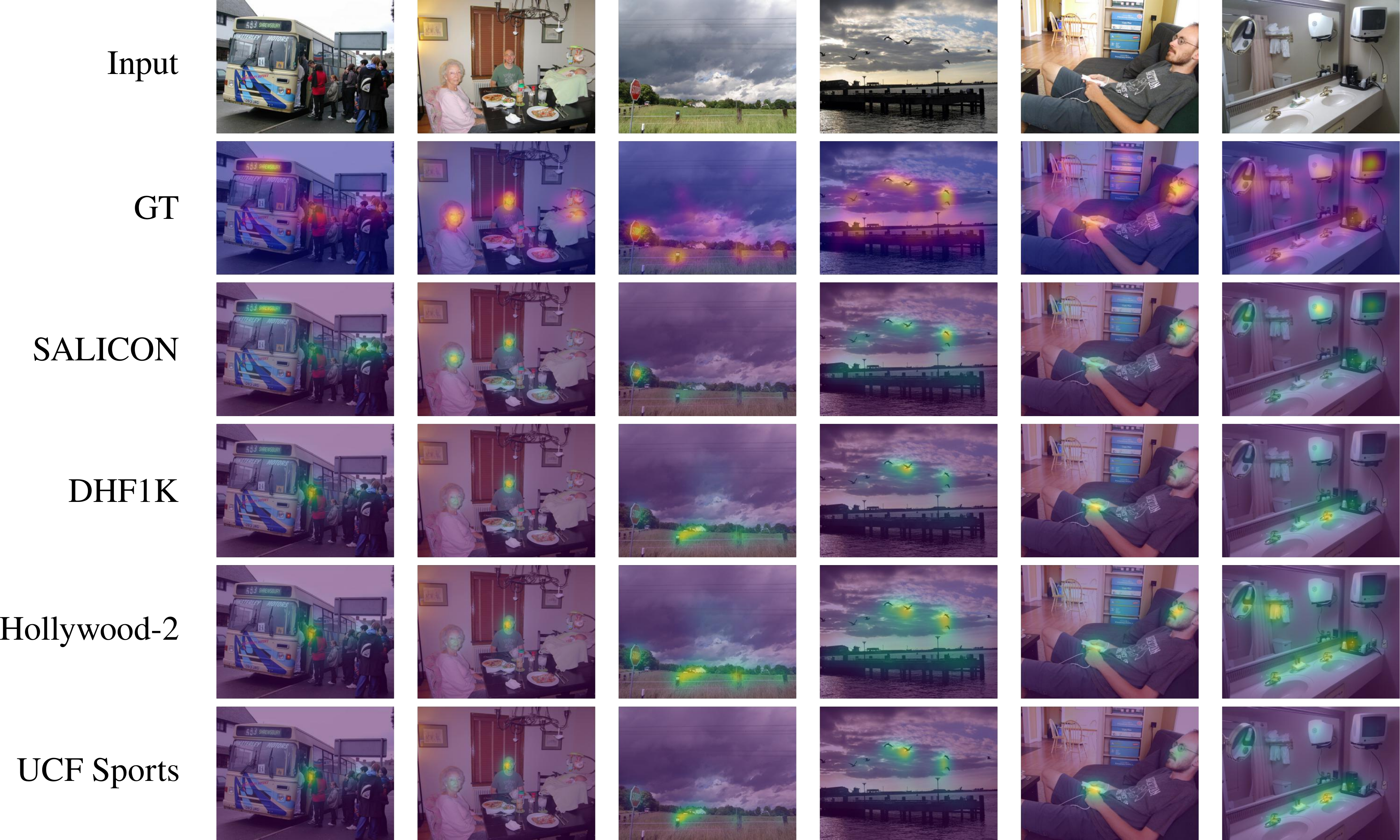}
\caption{
Cross-domain predictions for SALICON.
The images shown are drawn from the SALICON validation set.
The predictions are generated with the same trained UNISAL model, but different domain-adaptive settings.
The leftmost column shows the dataset whose modules were selected for the corresponding row.
}
\label{fig:crs_SALICON}
\end{figure*}

\begin{figure*}[t]
\centering
\includegraphics[width=\textwidth]{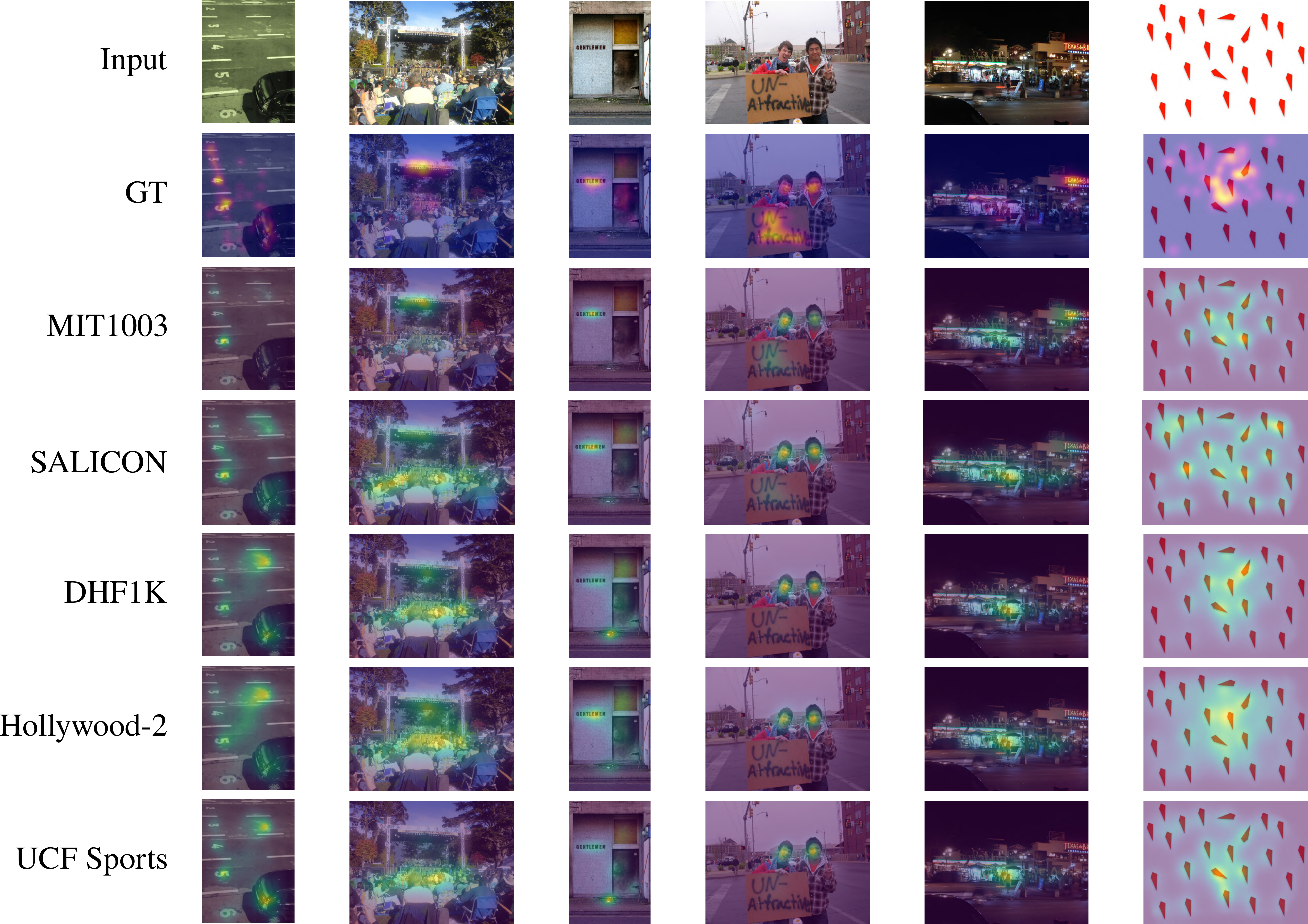}
\caption{
Cross-domain predictions for MIT1003.
The images shown are drawn from the MIT1003 dataset.
The predictions are generated with the same trained UNISAL model, but different domain-adaptive settings.
The leftmost column shows the dataset whose modules were selected for the corresponding row.
MIT1003 denotes the SALICON-specific setting which was fine-tuned on MIT1003 samples.
}
\label{fig:crs_MIT}
\end{figure*}

Here, we analyze the impact of the domain-adaptive modules when predicting visual saliency on the same input.
Results for video saliency prediction are shown in the second part of the attached video file \emph{3601-supp.mp4} (also available at \url{https://www.youtube.com/watch?v=4CqMPDI6BqE}).
Figure~\ref{fig:crs_SALICON} and Figure~\ref{fig:crs_MIT} show the results for image saliency prediction on SALICON and MIT1003 data, respectively.
It is visible in Figure~\ref{fig:crs_SALICON} that the video-specific settings (DHF1K, Hollywood-2, UCF Sports) cause the model to focus less on text and to focus on a single central object compared to the SALICON-specific setting.
Similar observations can be made for the results shown in Figure~\ref{fig:crs_MIT}.

\begin{figure*}[t]
\centering
\includegraphics[width=\textwidth]{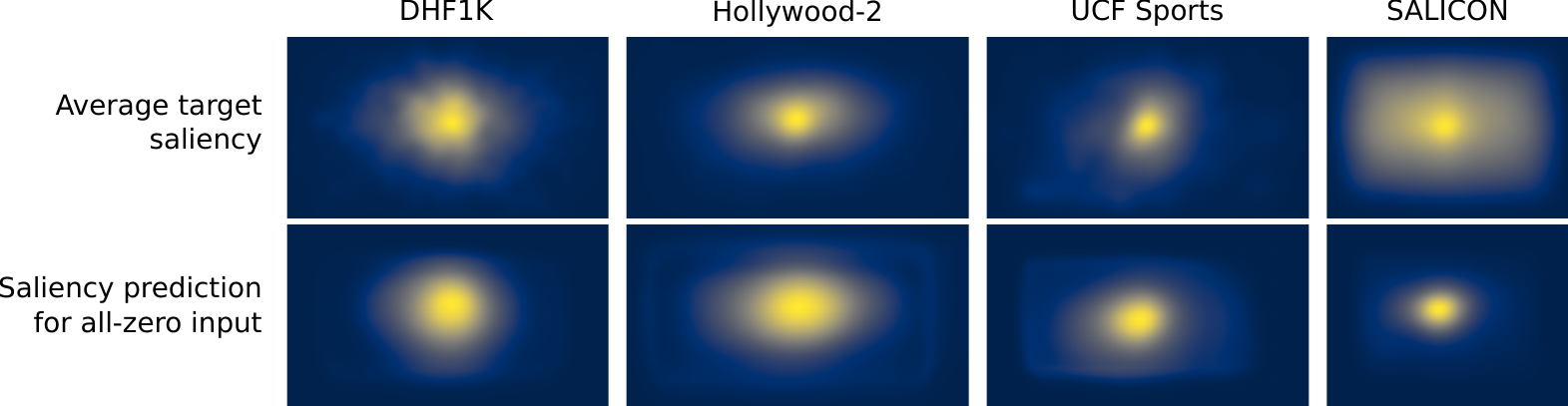}
\caption{
Saliency targets center biases vs.\ learned biases.
The upper row shows the average across all target training saliency maps for each dataset.
The lower row shows the prediction of the model for an all-zero input, for different domain-adaptive settings.
}
\label{fig:biases}
\end{figure*}

\section{Additional Center Bias Analysis}\label{Bias}
Here, we aim to evaluate the ability of the domain-adaptive learned Gaussian prior maps to capture the dataset-specific center biases.
The results are shown in Figure~\ref{fig:biases}.
The upper row shows the averaged saliency targets for each training dataset as an approximation of the true center biases.
In order to reveal the learned center biases, saliency predictions based on an all-zero input are generated for each set of domain-adaptive modules.
For the video saliency datasets, the learned bias reflects the true biases visibly well.
For SALICON, the true bias is significantly wider than the learned bias.
A possible explanation is that the spread-out true bias for SALICON is not caused by a more spread-out center bias of the viewers, but rather by a spread-out placement of salient objects.

\begin{table*}[t]
\caption{
Ablation study of the domain-adaptive modules on the DHF1K and SALICON validation sets.
The proposed components are added individually to a new baseline (\emph{Baseline+...+Smoothing}) to quantify their contribution.
Training setting (vi) is used for this study.
}
\label{tab:ablation}\centering
\resizebox{\textwidth}{!}{%
\huge
\begin{tabular}{@{}l|ccccc|ccccc@{}}
\toprule
\multirow{2}{*}{\diagbox{Config.}{Dataset}} & \multicolumn{5}{c|}{DHF1K} & \multicolumn{5}{c}{SALICON} \\
\cmidrule(l){2-11}
& KLD $\downarrow$ & AUC-J $\uparrow$ & SIM $\uparrow$ & CC $\uparrow$ & NSS $\uparrow$ & KLD $\downarrow$ & AUC-J $\uparrow$ & SIM $\uparrow$ & CC $\uparrow$ & NSS $\uparrow$ \\ \midrule
Baseline + ... + Smoothing* & 1.770 & 0.882 & 0.295 & 0.416 & 2.305 & 0.369 & 0.848 & 0.690 & 0.799 & 1.654 \\
* + DABN & 1.852 & 0.880 & 0.317 & 0.396 & 2.212 & 0.355 & 0.851 & 0.717 & 0.807 & 1.747 \\
* + DA-Gaussians & 1.748 & 0.884 & 0.315 & 0.412 & 2.278 & 0.386 & 0.848 & 0.679 & 0.794 & 1.647 \\
* + DA-Fusion & 1.706 & 0.888 & 0.326 & 0.434 & 2.437 & 0.326 & 0.854 & 0.712 & 0.820 & 1.750 \\
* + DA-Smoothing & 1.754 & 0.883 & 0.304 & 0.418 & 2.302 & 0.379 & 0.847 & 0.683 & 0.793 & 1.677 \\
* + BypassRNN & 1.784 & 0.882 & 0.322 & 0.412 & 2.302 & 0.356 & 0.853 & 0.695 & 0.819 & 1.721 \\
\bottomrule
\end{tabular}%
}
\end{table*}

\section{Additional Ablation Studies}
\label{Abla}
In the main paper, we perform an ablation study on the components of the proposed methods.
Here, we further ablate the individual domain-adaptive modules in Table~\ref{tab:ablation}.
We use the same evaluation metrics as in the main paper and perform the study on the DHF1K and SALICON datasets.
As a baseline for this study we use the \emph{Baseline} model of the main ablation study with modules added up to and including the \emph{Smoothing} module.
Then we add the individual domain-adaptive modules to this new baseline to analyze their respective effectiveness.
Specifically, we add the domain-adaptive batch normalization (\emph{DABN}), Gaussians (\emph{DA-Gaussians}), Fusion (\emph{DA-Fusion}), Smoothing (\emph{DA-Smoothing}), and the Bypass RNN (\emph{BypassRNN}).
The results in Table~\ref{tab:ablation} show that each domain-adaptive module contributes differently to the performance, in which the DA-Fusion contributes the most for both dynamic and static scenes.
This is consistent with our analyses in the main paper which indicate that this module has an important contribution towards mitigating the domain shift.

\begin{table*}[t]
\caption{
Cross-dataset generalization analysis of the UNISAL model on the SALICON benchmark test set.
The training settings (i) to (vi) denote training with: (i) DHF1K, (ii) Hollywood-2, (iii) UCF Sports, (iv) SALICON, (v) DHF1K+Hollywood-2+UCF Sports, and (vi) DHF1K+Hollywood-2+UCF Sports+SALICON.
}
\label{tab:benchS}
\centering
\begin{tabular}{@{}l|ccccccc@{}}
\toprule
& KLD$\downarrow$  & AUC-J $\uparrow$ & SIM $\uparrow$  & s-AUC $\uparrow$  & CC $\uparrow$  & NSS $\uparrow$  & IG $\uparrow$ \\ \midrule
Training setting (i) & 0.45 & 0.83 & 0.65 & 0.67 & 0.75 & 1.61 & 0.43 \\
Training setting (ii) & 0.50 & 0.83 & 0.63 & 0.67 & 0.73 & 1.52 & 0.35 \\
Training setting (iii) & 0.82 & 0.81 & 0.61 & 0.67 & 0.65 & 1.42 & 0.00 \\
Training setting (iv) & 0.42 & 0.86 & 0.78 & 0.74 & 0.88 & 1.95 & 0.72 \\
Training setting (v) & 0.48 & 0.83 & 0.66 & 0.66 & 0.74 & 1.61 & 0.44 \\
Training setting (vi) & 0.35 & {0.86} & {0.78} & {0.74} & {0.88} & {1.95} & 0.78 \\ \bottomrule
\end{tabular}%
\end{table*}

\section{SALICON Cross-Dataset Generalization}
\label{SALICON}
Here we analyze the cross-dataset generalization of the proposed UNISAL model for image saliency prediction on the SALICON benchmark test set.
Specifically, we analyze the performance of our UNISAL model on the SALICON dataset when training with different datasets, \ie, the six training settings described in the main paper, where setting (vi) is our final model.
In this study, we follow the standard SALICON benchmark evaluation pipeline and include two additional metrics of KL-divergence (\emph{KLD}) and Information Gain (\emph{IG}).
The results are shown in Table~\ref{tab:benchS}.
We observe that the model performs slightly worse when training on video datasets only compared to training on SALICON, even when jointly training with the three video datasets (setting (v)).
This observation confirms the existence of a domain shift between image and video saliency data.
On the other hand, when jointly training with video and image datasets, the performance is boosted on some metrics while remaining stable on the others.
This further validates the effectiveness of the proposed UNISAL approach to unify video and image saliency modeling.

\begin{table*}[t]
\caption{
Results on the MIT300 benchmark with probabilistic predictions (see Section~\ref{MIT300Prob}) with training setting (vi).
}
\label{tab:MITProb}
\centering
\begin{tabular}{@{}l|ccccccc@{}}
\toprule
& KLD$\downarrow$  & AUC-J $\uparrow$ & SIM $\uparrow$  & s-AUC $\uparrow$  & CC $\uparrow$  & NSS $\uparrow$  & IG $\uparrow$ \\ \midrule
UNISAL & 0.415 & 0.877 & 0.675 & 0.784 & 0.785 & 2.369 & 0.951 \\
\bottomrule
\end{tabular}%
\end{table*}

\section{MIT300 Probabilistic Benchmark Results}
\label{MIT300Prob}
All aforementioned results in this paper are computed without adapting the predicted saliency maps for individual evaluation metrics.
This is common practice and ensures comparability.
However, recent research \cite{Kummerer_2018_ECCV} has pointed out that the metrics are mutually inconsistent and that one set of saliency maps cannot perform equally well on all metrics.
For example, the \emph{s-AUC} metric requires division with the center bias map and the CC and \emph{KLD} metrics require smoothing.
Therefore, the MIT300 benchmark offers the possibility to submit probabilistic predicted saliency maps that are then mathematically optimized for each metric separately.
The results of the proposed UNISAL model on the probabilistic MIT300 benchmark are shown in Table~\ref{tab:MITProb}.

\section{Details for Quantitative Evaluation}
\label{Scoring}
\subsection{Scoring SalEMA with Training Setting (vi)}
For fairness of comparison, we score the SalEMA model\cite{Linardos2019} after fine-tuning it with training setting (vi), \ie, DHF1K+Hollywood-2+UCF Sports+SALICON.
For this, we use the official implementation provided by the authors under \url{https://github.com/Linardos/SalEMA/}.
We fine-tune the \emph{SalEMA30.pt} weights with the default training settings.
SALICON images are treated as single-frame videos.
The scores are computed on the test sets of UCF Sports and Hollywood-2 and the validation sets of DHF1K and SALICON, whose test sets are held-out for benchmarking.

\subsection{Scoring ACLNet on SALICON}
To obtain an additional baseline for image saliency prediction performance of an existing video saliency model besides SalEMA, we score the ACLNet model on the SALICON validation set (the test set is held-out for benchmarking).
We compute the scores when using either the auxiliary image saliency prediction output or the LSTM output of the model.
We find that the scores of the auxiliary output are better for all metrics and consequently report these in the paper.

\subsection{Sources of Other Benchmark Scores}
The scores of previous video saliency models on the DHF1K, UCF-Sports and Hollywood-2 datasets are obtained from \cite{wang2019revisiting}.
The scores of the previous image saliency models on the SALICON and MIT300 benchmarks were obtained from the respective papers.

\subsection{Generating MIT300 predictions}
As suggested by the benchmark authors, we fine-tune the model on the MIT1003 dataset before generating the MIT300 predictions.
Similar to \cite{kummerer2016deepgaze}, we fine-tune on MIT1003 with 10-fold cross validation.
The MIT300 predictions are then generated by averaging the log-probabilities of the 10 fine-tuned models.

\section{Code}\label{Code}
The full code for evaluating and training the UNISAL model is available at \url{https://github.com/rdroste/unisal}.

{\small
\bibliographystyle{splncs04}
\bibliography{mybib.bib}
}